\documentclass{OAGM}
\OAGMarXiv{1304.1876}

\usepackage{graphics}
\usepackage{pifont}
\usepackage{times}
\usepackage{epsfig}
\usepackage{graphicx}
\usepackage{amsmath}
\usepackage{amssymb}
\usepackage{tabularx} 
\usepackage[small]{caption}
\usepackage{pifont}
\usepackage[latin1]{inputenc}
\usepackage{multirow} 
\usepackage{rotating}

\newcommand{\npropto}{%
 \propto\hskip-2ex/\hskip1ex
}

\newcommand{\err}{%
 \scriptsize\ding{53}\normalsize
}
\newcommand{\ok}{%
 \scriptsize\ding{51}\normalsize
}

\newcommand{\eg}{%
 e.g.\
}

\title{Euclidean Upgrade from a Minimal Number of Segments}

\author{Tanja Schilling \and Tom\'{a}\v{s} Pajdla\\
Center for Machine Perception, Department of Cybernetics\\
Faculty of Electrical Engineering, Czech Technical University in Prague}

\begin{document}
\maketitle

\begin{abstract}
In this paper, we propose an algebraic approach to upgrade a projective reconstruction to a Euclidean one, and aim at computing the rectifying homography from a minimal number of 9 segments of known length. Constraints are derived from these segments which yield a set of polynomial equations that we solve by means of Gröbner bases. We explain how a solver for such a system of equations can be constructed from simplified template data. Moreover, we present experiments that demonstrate that the given problem can be solved in this way.
\end{abstract}

\section{Introduction}\label{sec:intro}

A projective reconstruction can be computed from image correspondences alone, without any information about calibration or pose of the cameras. But additional knowledge about the scene is required to subsequently recover Euclidean structure~\cite{Hartley.Zisserman_MultipleViewGeometry_2006}. The problem of computing such a homography that upgrades a projective reconstruction to a metric one has been treated in many publications, \eg  in~\cite{Ponce_ComputingMetricUpgrades_2001,Heyden.Astrom_EuclideanReconstructionfrom_1997}.

In our case, the additional knowledge about the scene comprises a set of segments with known lengths, i.e.\ pairs of points with known distances between them. This problem has already been addressed in~\cite{Liebowitz.Carlsson_UncalibratedMotionCapture_2003}, where parallel projection was assumed and hence an affine reconstruction was supposed to be known.

The projective case was considered in~\cite{Ronda.Valdes_EuclideanUpgradingfrom_2010}. The authors introduced the concept of a Quadric of Segments (QoS), defined in a higher-dimensional space by the set of segments of known length, that can be computed linearly. Euclidean structure can be recovered from the QoS in closed form, exploiting the fact that all spheres intersect the plane at infinity in the absolute conic~\cite{Hartley.Zisserman_MultipleViewGeometry_2006}. Although the linear computability is certainly advantageous, the high number of segments (54 in the 3D case) required to define the QoS constitutes a main drawback. Given that not many segments of known length are available in various conceivable settings, we therefore seek a solution that requires only as few segments as possible.

The constraints on the upgrading homography derived from these segments constitute a system of non-linear algebraic equations. Polynomial systems occur in various computer vision problems and many of them have been solved by means of Gröbner bases~\cite{Bujnak.etal_3DReconstructionfrom_2009,Byroed.etal_FastOptimalThree_2007,Stewenius2005}. But there is no easy, straightforward method to solve general polynomial systems efficiently and robustly. Instead, each particular problem usually requires the manual design of a suitable Gröbner basis solver. However, an automatic generator of minimal problem solvers was presented in~\cite{Kukelova.etal_AutomaticGeneratorMinimal_2008}. 

In this paper, we show that it is possible to compute the homography that upgrades a projective reconstruction to a Euclidean one algebraically from only 9 segments of known length. The following section states the problem in detail. Section~\ref{sec:groebner} briefly introduces the notion of Gröbner bases, on which our approach is based. We explain how a special solver for the problem at hand can be constructed and applied in section~\ref{sec:solver}. Finally, experimental results are presented in section~\ref{sec:experiments}.

\section{Problem statement}\label{sec:problem}

The image projection of a scene point $X_i$, represented by its homogeneous coordinates~\cite{Hartley.Zisserman_MultipleViewGeometry_2006}, is denoted as	$x_i \propto \mathbf{P} X_i$, i.e.\ an $\alpha_i \in \mathbb{R}$ exists, such that $x_i = \alpha_i\mathbf{P} X_i$. Let us assume that we have measured $n$ image coordinates $\hat{x}_i \npropto x_i$, $i \in \mathbb{N}$, and computed a projective reconstruction ($\mathbf{\hat{P}}$, \{$\hat{X}_i$\}), where $\mathbf{\hat{P}} \npropto  \mathbf{P}$ and $\hat{X}_i \npropto X_i$, from those points, such that $\hat{x}_i \propto \mathbf{\hat{P}} \hat{X}_i$. Aiming at recovering Euclidean structure from this, we have to find a non-singular matrix $\mathbf{H}\in \mathbb{R}^{4 \times 4}$ which upgrades $\hat{X}_i$ and $\mathbf{\hat{P}}$ by a projective transformation, such that $\mathbf{P} X_i \propto \mathbf{\hat{P}} \mathbf{H}^{-1} \mathbf{H} \hat{X}_i$ and $X_i \propto \mathbf{H} \hat{X}_i$.

In order to reduce the number of unknowns in $\mathbf{H}$, we fix the reference frame by choosing three points $\hat{X}_i$ that determine the origin, the $x$-axis and the $xy$-plane. From now on, let us assume that all points $\hat{X}_i$ have already been mapped by a suitable similarity transform such that there exist three points $(0,0,0,1)^{\top}$, $(\hat{x}_i,0,0,1)^{\top}$ and $(\hat{x}_j,\hat{y}_j,0,1)^{\top}$, $\hat{x}_i,\hat{x}_j,\hat{y}_j \in \mathbb{R}\setminus\left\{0\right\}$, which determine the coordinate frame.

Mapping the point of origin to itself, $(0,0,0,1)^{\top} \propto \mathbf{H} (0,0,0,1)^{\top}$, points on the $x$-axis to the $x$-axis, $(x_i,0,0,1)^{\top} \propto \mathbf{H} (\hat{x}_i,0,0,1)^{\top}$, and points in the $xy$-plane to the $xy$-plane again, $(x_j, y_j,0,1)^{\top} \propto \mathbf{H} (\hat{x}_j,\hat{y}_j,0,1)^{\top}$, the transformation matrix has to be of the form
\begin{equation}\label{eq:H}
\mathbf{H} =  
\left( \begin{array}{cccc}
 h_{1} & h_{2} & h_{3} & 0 \\
 0 & h_{4} & h_{5} & 0 \\
 0 & 0 & h_{6} & 0 \\
 h_{1}-h_{9} & h_{7} & h_{8} & h_{9} \end{array} \right). 
 \end{equation} 
 
Two constraints on $\mathbf{H}$ easily arise from that. First, the projection matrix has to be invertible and therefore $\mathbf{H}$ must fulfill
\begin{equation}\label{eq:ineq-det}
0 \neq \det(\mathbf{H}) = h_{1}h_{4}h_{6}h_{9}.
\end{equation}

Secondly, we fix the scale of $\mathbf{H}$ by demanding that for one point $\hat{X}_i$, $1\leq i\leq n$
\begin{equation}\label{eq:ineq-inf}
1 = \mathbf{H}_4\hat{X}_i = X_{4i}
\end{equation} 
with $\mathbf{H}_{4}$ denoting the $4$-th row of $\mathbf{H}$ and $X_{4i}$ being the 4-th element of $X_i$.

Furthermore, we assume that there are $N$ pairs of points, $(X_{i},Y_{i})$ which represent segments of known lengths $d_i$, such that $\left\| X_{i} - Y_{i} \right\| = d_i$ for all $i=1\ldots N$. Replacing $X_i$ by $\mathbf{H}\hat{X}_i$, yields the following constraint on $\mathbf{H}$: 
\begin{equation}\label{eq:constraint}
f_{i}(\mathbf{h}) = 0 = \sum_{l=1}^3 (\mathbf{H}_{l}\hat{X}_{i}\mathbf{H}_{4}\hat{Y}_{i}-\mathbf{H}_{4}\hat{X}_{i}\mathbf{H}_{l}\hat{Y}_{i})^2 - (\mathbf{H}_{4}\hat{X}_{i}\mathbf{H}_{4}\hat{Y}_{i})^2 d_{i}^2 
\end{equation}
where $\mathbf{H}_{l}$ denotes the $l$-th row of $\mathbf{H}$.

Equation~\eqref{eq:constraint} constitutes a homogeneous polynomial of degree 4 in 9 variables, with 97 terms in the general case. Therefore, at least 9 segments $(X_{i},Y_{i})$ are required to obtain the 9 equations, which determine $\mathbf{H}$. Introducing an additional variable $h_{10}$, we can rewrite the inequality~\eqref{eq:ineq-det} as equality~\cite{Becker.Weispfenning_GroebnerBases:Computational_1993}
\begin{equation}\label{eq:det}
		0 = 1 - h_{1}h_{4}h_{6}h_{9}h_{10}, 
\end{equation}
as well as
\begin{equation}		
	\label{eq:inf}	0 = 1 - \mathbf{H}_4 \hat{X}_{i}.
\end{equation}

Now, the problem is to solve the system of $m = N+2$ algebraic equations, $N \geq 9$, in 10 variables $h_1,\ldots,h_{10}$,
\begin{equation}
0 = f_1(\mathbf{h}) = \cdots = f_m(\mathbf{h}).
\end{equation}

\section{Gröbner bases}\label{sec:groebner}

Systems of polynomial equations can be solved efficiently by means of Gröbner bases~\cite{Cox.etal_IdealsVarietiesand_1992}. $F$ denotes the set of $m$ polynomials $F = \left\{ f_1(\mathbf{h}), \ldots ,f_m(\mathbf{h}) | f_i(\mathbf{h}) \in K[h_1, \ldots , h_n] \right\}$ in $n$ variables $\mathbf{h} = (h_1, \ldots , h_n)$ over a field $K$. The ideal $I = \left\langle F \right\rangle$ generated by $F$ is the set of all polynomial linear combinations 
\begin{equation}
I = \left\{ \sum^m_{i=1} f_i(\mathbf{h})q_i(\mathbf{h}) | q_i(\mathbf{h}) \in K[h_1, \ldots , h_n] \right\}. 
\end{equation}

A Gröbner basis is a special set of generators with desirable algorithmic properties. In particular, a Gröbner basis of an ideal $I$ has the same set of solutions as $I$. But similar to a system of linear equations after Gaussian elimination, the solutions of $I$ can be easily identified in the corresponding Gröbner basis w.r.t.\ a lexicographical monomial ordering~\cite{Cox.etal_IdealsVarietiesand_1992}.

Theoretically, the Gröbner basis can be computed from any generating set of $I$ by a method called Buchberger's algorithm~\cite{Cox.etal_IdealsVarietiesand_1992}.
The basic mechanism is to take each pair $(f_i(\mathbf{h}), f_j(\mathbf{h}))$ from $F$, $f_i(\mathbf{h}) \neq f_j(\mathbf{h})$, compute its $S$-polynomial (see appendix), reduce it by $F$ and add the remainder to $F$ if it is not zero. This is done until the $S$-polynomials of all pairs in $F$ reduce to zero.
 
This problem is known to be EXPSPACE-complete in general~\cite{Kuehnle.Mayr_ExponentialSpaceComputation_1996}. Nevertheless, much better bounds can be found for many cases that actually occur in practice and several well-known methods exist to improve the basic algorithm. 

However, computing a Gröbner basis straightaway from the set of equations introduced in section~\ref{sec:problem} with floating point arithmetic is not practicable for two reasons: One obstacle is that even with improved versions of Buchberger's algorithm many $S$-polynomials are constructed in vain as they finally reduce to zero and hence do not contribute to the final basis, merely slowing the entire Gröbner basis computation down. Another difficulty arises as a result of accumulating round-off errors in floating point arithmetic during repeated reductions of $S$-polynomials. Except for extremely simple examples, these round-off errors make it impossible to decide whether a particular coefficient very close to zero should be considered as zero causing the cancellation of the corresponding term or not. Thus a special solver that is adapted to the given problem has to be created.

\section{Solver design and application}\label{sec:solver}

We build our solver by computing a Gröbner basis of a template system of equations first. These polynomials are generated as explained in section~\ref{sec:problem}, but originate from a simplified set of segments $(X_i, Y_i)$ with integer coordinates. Calculations are done in a finite field $\mathbb{Z}_p$. During the computation, we record which of the pairs of polynomials taken from the intermediate basis form $S$-polynomials that do not reduce to zero during the process. In so doing, we get a computation template that contains only those pairs of polynomials that actually contribute to the final Gröbner basis. In principle, this step has to be performed only once.

Afterwards, we can apply that solver to compute a Gröbner basis from a general polynomial system in floating point arithmetic, taking only the previously recorded pairs of polynomials into account, which speeds up this procedure. To avoid the above mentioned problematic effects of round-off errors, we could have memorized when which integer coefficient during the template computation gets zero as proposed in~\cite{Traverso.Zanoni_NumericalStabilityand_2002} and proceed accordingly. But since storing and reading this information takes a considerable amount of memory and time, we favor the more practicable way of simply processing the template data once again, simultaneously with the polynomial system that we want to solve. Whenever a coefficient of the template data (computed in $\mathbb{Z}_p$) becomes zero, the corresponding floating point coefficient is set to zero, too.
 
We assume that the sequence of operations to construct the Gröbner basis is basically identical for different sets of polynomials, given that the equations in those sets contain the same monomials and differ only in their coefficients~\cite{Traverso_GroebnerTraceAlgorithms_1988}. With this assumption, we rely on the fact that Buchberger's algorithm and its improved variants do not consider the values of non-zero coefficients for the choice of critical pairs, the detection of unnecessary pairs or the selection of reductors.

The crucial point in this scheme is to find an appropriate template system that is simple enough to be feasible yet general enough to be used for the original problem. The difficulty is in the fact that having identical monomials in the template polynomials and in the original is required to achieve the same sequence of computation but does not necessarily lead to the desired result.

To generate the template set, we simplified the general problem by using small integers coordinates for all segments $\left\|X_{i} - Y_{i}\right\| = d_i$, such that also $d_i \in \mathbb{Z}\setminus\left\{0\right\}$. That means each segment has to fulfill 
\begin{equation}\label{eq:pythquad}
d^2 = \left\|X_{i} - Y_{i}\right\|^2 = a^2 + b^2 + c^2,
\end{equation}
where $a,b,c,d \in \mathbb{Z}$, $d\neq0$, thus forming a Pythagorean quadruple if $a,b,c \in \mathbb{Z}\setminus\left\{0\right\}$ or a Pythagorean triple respectively if $a=0$, $b\neq0$ and $c\neq0$. The possibility to use such a special polynomial system as a template for the general problem is justified by the fact that every triple in $\mathbb{R}^3$ has a sufficiently precise scaled representation as a Pythagorean triple in $\mathbb{Z}^3$ ~\cite{Shiu_ShapesandSizes_1983}. Hence, we are looking for a generic Pythagorean case which is feasible to compute and at the same time implementable for a wide range of practically occurring systems originating from real coefficients. 

\section{Experiments}\label{sec:experiments}

\subsection{Solver generation}\label{sec:experiments.construct}

Although we aim to generate a solver to compute $\mathbf{H}$ from the minimal number of segments $N=9$, we conducted experiments to assess the influence of the number of segments to the presented method.

The template data set from which the solver is built consists of $N$ pairs of points $\left(X_{i}, Y_{i}\right)$ that are generated from Pythagorean quadruples as outlined above, and a homography $\mathbf{H}$. More precisely, $N-1$ Pythagorean quadruples $\left(a_{i}, b_{i}, c_{i}, d_{i}\right)$, where $0<a_{i}, b_{i}, c_{i}, d_{i}<50$, are selected randomly, such that $\left(a_{i}, b_{i}, c_{i}, d_{i}\right)\neq\left(sa_{j}, sb_{j}, sc_{j}, sd_{j}\right)$ for any scaling factor $s\in \mathbb{Z}$, for all $i,j=1\ldots N$ and $i\neq j$. Euclidean coordinates of $X_{i}$ are chosen randomly as well, whereupon the points $Y_{i}$ are calculated, such that
\begin{equation}
X_{i}-Y_{i} = (X_{1i},X_{2i},X_{3i},1)^{\top}-(Y_{1i},Y_{2i},Y_{3i},1)^{\top} = (a_{i}, b_{i}, c_{i},0)^{\top}.
\end{equation}

In general, $X_{i}$ and $Y_{i}$ coordinates are non-zero integers, $0<\left|X_{ji}\right|,\left|Y_{ji}\right|<100$, except for the first 2 segments
\begin{equation}\begin{aligned}
X_{1}&=(0,0,0,1)^{\top}, 						& Y_{1}&=(Y_{11},0,0,1)^{\top},\\ 
X_{2}&=(X_{12},X_{22},0,1)^{\top}, 	& Y_{2}&=(Y_{12},Y_{22},Y_{32},1)^{\top},
\end{aligned}\end{equation}
which are required to determine the $x$-axis and the $xy$-plane as explained in section~\ref{sec:problem}. $\mathbf{H}$ is generated according to equation~\eqref{eq:H}, with random non-zero integer elements $0<h_{k}<20$ for all $k=1\ldots9$.

The resulting $N$ pairs of distorted points 
\begin{equation}
\left(\mathbf{H}^{-1}X_{i}, \mathbf{H}^{-1}Y_{i}\right) = \left(\hat{X}_{i}, \hat{Y}_{i}\right) 
\end{equation}
yield $N+2$ polynomials according to equations~\eqref{eq:constraint}, \eqref{eq:det} and \eqref{eq:inf}. A Gröbner basis is computed from this system of equations in a finite field $\mathbb{Z}_p$ with $p=332251314113$ as this proved to be a sufficiently large prime number in earlier experiments. In that way, solvers are built for 20 different template data sets for each of 10 distinct values of $N$.  

As outlined in section~\ref{sec:groebner}, the intermediate basis $F$ grows until the $S$-polynomials of all pairs $(f_i(\mathbf{h}), f_j(\mathbf{h}))$ in $F$ reduce to zero, usually producing a large final Gröbner basis in this way. Therefore, its reduced Gröbner basis~\cite{Cox.etal_IdealsVarietiesand_1992} is computed afterwards, from which the solution can be easily obtained and which is unique for the considered ideal and a given monomial ordering. 

The reduced Gröbner basis for the ideal generated by the considered polynomial system derived from $N$ segments consists of the following 13 simple equations $g_{i}$, that only differ in the coefficients $c_{ij}$.
\begin{equation}\label{eq:g1}\begin{aligned}
g_{1} &= 0  = c_{1,1}h_{1}+c_{1,2}   &	 g_{2} &= 0  = c_{2,1}h_{2}+c_{2,2}  	\\
g_{3} &= 0  = c_{3,1}h_{3}+c_{3,2}   							&	 g_{4} &= 0  = c_{4,1}h_{7}+c_{4,2}  	\\
g_{5} &= 0  = c_{5,1}h_{8}+c_{5,2}    						&	 g_{6} &= 0  = c_{6,1}h_{9}+c_{6,2} 	\\
g_{7} &= 0  = c_{7,1}h_{4}+c_{7,2}h_{5}    				&	 g_{8} &= 0  = c_{8,1}h_{5}^2+c_{8,2} \\
g_{9} &= 0  = c_{9,1}h_{6}^2+c_{9,2}   					 	&	g_{10} &= 0  = c_{10,1}h_{10}^2+c_{10,2} \\
g_{11} &= 0  = c_{11,1}h_{5}h_{6}+c_{11,2}h_{10} 	&	g_{12} &= 0  = c_{12,1}h_{5}h_{10}+c_{12,2}h_{6} \\
g_{13} &= 0  = c_{13,1}h_{6}h_{10}+c_{13,2}h_{5}
\end{aligned}\end{equation}

These equations yield 4 solutions for $\mathbf{H}$, varying in the signs of $h_4$, $h_5$, and $h_6$, that maps all points $\hat{X}_{i}$ either to 
\begin{equation}\begin{aligned} 
 X_i^{(++)} &= (X_{1i},X_{2i},X_{3i},1)^{\top}, & X_i^{(-+)} &= (X_{1i},-X_{2i},X_{3i},1)^{\top},\\
 X_i^{(+-)} &= (X_{1i},X_{2i},-X_{3i},1)^{\top}\ \ \mbox{or} & X_i^{(--)} &= (X_{1i},-X_{2i},-X_{3i},1)^{\top}, 
\end{aligned}\end{equation} and $\hat{Y}_{i}$ analogously, depending on whether both $h_5$ and $h_6$ are positive, $h_5<0$ and $h_6>0$, $h_5>0$ and $h_6<0$ or both are negative. Hence, a $180^\circ$ rotation of all segments ($X_i^{(++)},Y_i^{(++)}$) about the $x$-axis corresponds to the set of segments \{($X_i^{(--)},Y_i^{(--)}$)\}, and mirroring these sets on the $xz$-plane yields the other two sets of segments \{($X_i^{(-+)},Y_i^{(-+)}$)\} and \{($X_i^{(+-)},Y_i^{(+-)}$)\}.     

In contrast to the explained basic Buchberger's algorithm, sets of pairs of polynomials $\left\{(f_i(\mathbf{h}), f_j(\mathbf{h}))\right\}$ instead of single pairs are reduced simultaneously during so called multi-reduction steps in the modified version~\cite{Brickenstein_Slimgb:GroebnerBases_2005} of the algorithm that we use in our experiments. 

The so computed reduced Gröbner bases from various template data sets and different values of $N$ vary only in the coefficients $c_{ij}$ of equations~\eqref{eq:g1}. Table~\ref{fig:results.construct} shows that the Gröbner basis computation is faster for higher number of segments in terms of required multi-reduction steps and computing time. But more than 50 segments do not speed it up further. Unsurprisingly, the final basis contains more equations if more multi-reduction steps were necessary to compute it.   
\begin{table}
\begin{tabular}{cc}
\begin{minipage}[t]{0.48\textwidth}\begin{center}\small
\begin{tabular}{|c|c|c|c|}
\hline
$N$	&	Basis &	Multi-			&	Time \\
  	&	size 	&	reductions	&	in msec	\\
\hline
9	&	500	&	28	&	5170	\\
10	&	424	&	24	&	3828	\\
12	&	364	&	19	&	759	\\
15	&	206	&	17	&	140	\\
20	&	105	&	17	&	11	\\
25	&	95	&	14	&	10	\\
50	&	67	&	12	&	5	\\
100	&	67	&	12	&	7	\\
200	&	67	&	13	&	20	\\
500	&	67	&	16	&	57	\\
\hline
\end{tabular}
\caption{\small Solver generation: Size of the computed (not yet reduced) Gröbner basis, required multi-reduction steps and computation time for different numbers of segments $N$.}\label{fig:results.construct}
\end{center}\end{minipage} 
&
\begin{minipage}[t]{0.45\textwidth}\begin{center}\small
\begin{tabular}{|c|c|c|}
\hline
$N$	& Precision &	Time \\
		& in bit		&	(min:sec.msec)	\\
\hline
9	&	1088	&	2:49.23	\\
10	&	512		&	1:24.16	\\
12	&	448		&	0:12.50	\\
15	&	384		&	0:02.59	\\
20	&	192	&		0:00.11	\\
25	&	256		&	0:00.10	\\
50	&	256		&	0:00.05	\\
100	&	256		&	0:00.07	\\
200	&	256		&	0:00.20	\\
500	&	256		&	0:00.57	\\
\hline
\end{tabular}
\caption{\small Solver testing: Required floating point precision and computation time to calculate the correct solution for different numbers of segments $N$.}\label{fig:results.test1}
\end{center}\end{minipage}
\end{tabular}
\end{table}

Next, the generated solvers are tested to find the minimally required precision for floating point arithmetic. More exactly, we apply each solver to its template data set, but compute its reduced Gröbner basis in floating point arithmetic this time as explained in section~\ref{sec:solver}. If the absolute values of the difference between the so computed segment length and the true segment length is smaller than $10^{-9}$, computed and true segment length are considered to be equal and the used precision is therefore sufficient. Our implementations make use of the {GNU} Multiple Precision Arithmetic Library~\cite{GMP} to handle high precision floating point arithmetic as well as large integers. 

The required precisions for different $N$ are given in table~\ref{fig:results.test1}. As expected, a higher precision is required for a lower number of segments, but using more than 50 segments does not decrease the necessary precision further. 

\subsection{Application to exact data}\label{sec:experiments.process}

Now, we apply the solver to various floating point data sets. Each data set comprises $N$ pairs $\left(X_{i}, Y_{i}\right)$, and a homography 
$\mathbf{H}$. The Euclidean coordinates of the points $X_{i}$ and $Y_{i}$ are in general randomly chosen within a cube of side lengths 10. All coordinates $X_{ji},Y_{ji} \in \mathbb{R}$ are chosen such that $10X_{ji},10Y_{ji}\in \mathbb{Z}$, i.e.\ only coordinates with at most one digit after the decimal point are considered. Furthermore, $0\leq X_{ji},Y_{ji}\leq 10$ for all $j=1\ldots 3$ and $X_{4i}=Y_{ji}=1$ for all $i=1\ldots N$. The true segment length $d_i=\left\|X_{i}-Y_{i}\right\|$ varies. 

$\mathbf{H}$ is chosen in a way that $\mathbf{H}^{-1}$ maps 3 vertices $V_{l}$ of the cube to $V_{l}+T_{l} \propto \mathbf{H}^{-1}V_{l}$, where $T$ is a random vector of independent zero-mean Gaussian components with typical deviation 1, and all $h_k \in \mathbb{R}$. The system of polynomial equation is obtained from $N$ pairs $\left(\hat{X}_{i}=\mathbf{H}^{-1}X_{i}, \hat{Y}_{i}=\mathbf{H}^{-1}Y_{i}\right)$ according to equations~\eqref{eq:constraint}, \eqref{eq:det} and \eqref{eq:inf}. For different values of $N$, each of the 20 solvers created in section~\ref{sec:experiments.construct} is applied to 20 different data sets, to compute the respective reduced Gröbner bases, and then the sought homographies $\mathbf{H}'$.  Subsequently, the distorted points $\hat{X}_{i}$ and are $\hat{Y}_{i}$ are upgraded to $X'_{i}=\mathbf{H}'\hat{X}_{i}$ and $Y'_{i}=\mathbf{H}'\hat{Y}_{i}$ and the segment lengths $d'_i=\left\|X'_{i}-Y'_{i}\right\|$ are computed.

The error is calculated as the standard deviation of the differences between the computed and true segment lengths divided by the average segment length  $\sigma(d_i-d'_i)/\mu(d_i)$. However, solvers may fail under certain conditions, such that no solution can be obtained. In table~\ref{fig:results.process}, the percentage of successful computed solutions and the mean errors of those solutions are summarized for different numbers of segments $N$.

In our experiments, the solvers failed for instance on 2 data sets due to the chosen $\mathbf{H}^{-1}$ and a pair $\left(X_{i}, Y_{i}\right)$, that caused one or more coefficients in the resulting polynomial to be zero. Hence, a floating point exception occurred when division by zero was attempted during reduction by this polynomial. However, this problem occurred only in less than 0.1\% of all polynomials that we generated from test data. In all cases, where the solver could be successfully applied to compute a solution, the upgraded segment lengths nearly equal the true length. 
\begin{table}[t]
\begin{tabular}{cc}
\begin{minipage}[b]{0.35\textwidth}\begin{center}\small
\begin{tabular}{|c|c|r@{.}l|}
\hline
$N$	& Success	&	\multicolumn{2}{|c|}{Mean}	\\
	  & rate 		&	\multicolumn{2}{|c|}{error}	\\
\hline
9	&	100\% 		&	7&5e-37	\\
10	&	100\%		&	3&4e-45	\\
12	&	100\%		&	4&1e-70	\\
15	&	100\%		&	3&4e-82	\\
20	&	100\%	  &	2&3e-129	\\
25	&	100\%		& 1&4e-137	\\
50	&	95\%		&	2&3e-139	\\
100	&	95\%		&	1&2e-141	\\
200	&	85\%		&	4&5e-144	\\
500	&	85\%		&	3&2e-149	\\
\hline
\end{tabular}
\caption{\small Solver application to exact data: Success rate and mean errors for different numbers of segments $N$.}\label{fig:results.process}
\begin{tabular}{|c|c|c|}
\hline
$N$	& Success	&	Mean	\\
	  & rate 		&	error	\\
\hline
25	&	4\%  & 0.26	\\
50	&	48\% & 0.11	\\
100	&	49\% & 0.11	\\
\hline
\end{tabular}
\caption{\small Solver application to noisy data, $\sigma = 0.001$: Success rate and mean errors for different numbers of segments $N$.}\label{fig:results.noise}
\end{center}\end{minipage} &
\begin{minipage}[b]{0.6\textwidth}\begin{center}\small
\begin{tabular}{|c|r@{ }r@{ }r@{ }r@{ }r@{ }r@{ }r@{ }r@{ }r@{ }r@{ }r@{ }r@{ }r@{ }r@{ }r@{ }r@{ }r@{ }r@{ }r@{ }r@{ }|r|}
\hline
& \multicolumn{20}{|c|}{20 different data sets} & $\Sigma$	\\
\hline
\multirow{19}{1mm}{\begin{sideways}{19 different solvers}\end{sideways}}	
& \err	& \ok	& \err	& \err	& \ok	& \err	& \err	& \ok	& \err	& \ok	& \ok	& \err	& \ok	& \err	& \ok	& \err	& \ok	& \err	& \ok	& \err	& 9 \\
& \err	& \ok	& \ok	& \err	& \err	& \err	& \err	& \ok	& \ok	& \ok	& \ok	& \err	& \ok	& \err	& \ok	& \err	& \ok	& \err	& \err	& \err	& 9 \\
& \err	& \ok	& \err	& \err	& \ok	& \err	& \err	& \ok	& \ok	& \ok	& \ok	& \err	& \ok	& \err	& \ok	& \err	& \ok	& \err	& \err	& \err	& 9 \\
& \err	& \ok	& \ok	& \err	& \ok	& \err	& \err	& \ok	& \ok	& \ok	& \ok	& \err	& \ok	& \err	& \ok	& \ok	& \ok	& \err	& \ok	& \err	& 12 \\
& \err	& \ok	& \ok	& \err	& \err	& \err	& \err	& \ok	& \ok	& \ok	& \ok	& \err	& \ok	& \err	& \ok	& \err	& \ok	& \err	& \err	& \err	& 9 \\
& \err	& \ok	& \ok	& \err	& \err	& \err	& \err	& \ok	& \ok	& \ok	& \ok	& \err	& \ok	& \err	& \ok	& \err	& \ok	& \err	& \err	& \err	& 9 \\
& \err	& \ok	& \ok	& \err	& \ok	& \err	& \err	& \ok	& \err	& \ok	& \ok	& \err	& \err	& \err	& \ok	& \err	& \ok	& \err	& \err	& \err	& 8 \\
& \err	& \ok	& \ok	& \err	& \ok	& \err	& \err	& \ok	& \ok	& \ok	& \ok	& \err	& \ok	& \err	& \ok	& \ok	& \ok	& \ok	& \err	& \err	& 12 \\
& \err	& \ok	& \err	& \err	& \err	& \err	& \err	& \ok	& \ok	& \ok	& \ok	& \err	& \ok	& \err	& \ok	& \err	& \ok	& \err	& \err	& \ok	& 9 \\
& \err	& \ok	& \err	& \err	& \ok	& \err	& \err	& \ok	& \ok	& \ok	& \ok	& \err	& \ok	& \err	& \ok	& \err	& \ok	& \err	& \err	& \err	& 9 \\
& \err	& \ok	& \err	& \err	& \err	& \err	& \err	& \ok	& \ok	& \ok	& \ok	& \err	& \err	& \err	& \ok	& \ok	& \ok	& \err	& \ok	& \err	& 9 \\
& \err	& \ok	& \ok	& \err	& \ok	& \err	& \err	& \ok	& \ok	& \ok	& \ok	& \err	& \err	& \err	& \ok	& \ok	& \ok	& \err	& \err	& \err	& 10 \\
& \ok	& \ok	& \err	& \err	& \err	& \err	& \err	& \ok	& \ok	& \ok	& \ok	& \err	& \err	& \ok	& \ok	& \ok	& \ok	& \err	& \err	& \ok	& 11 \\
& \err	& \ok	& \err	& \err	& \err	& \err	& \err	& \ok	& \ok	& \ok	& \ok	& \err	& \ok	& \err	& \ok	& \err	& \ok	& \err	& \ok	& \err	& 9 \\
& \err	& \ok	& \err	& \err	& \ok	& \err	& \err	& \ok	& \ok	& \ok	& \ok	& \err	& \ok	& \err	& \ok	& \err	& \ok	& \err	& \err	& \err	& 9 \\
& \err	& \ok	& \ok	& \err	& \ok	& \err	& \err	& \ok	& \ok	& \ok	& \ok	& \err	& \err	& \err	& \ok	& \err	& \ok	& \err	& \ok	& \err	& 10 \\
& \err	& \ok	& \err	& \err	& \err	& \err	& \err	& \ok	& \ok	& \ok	& \ok	& \err	& \ok	& \err	& \ok	& \ok	& \err	& \err	& \ok	& \err	& 9 \\
& \err	& \ok	& \ok	& \err	& \ok	& \err	& \err	& \ok	& \ok	& \ok	& \ok	& \err	& \ok	& \err	& \ok	& \err	& \err	& \err	& \ok	& \ok	& 11 \\
& \err	& \ok	& \ok	& \err	& \ok	& \err	& \err	& \ok	& \ok	& \ok	& \ok	& \err	& \ok	& \err	& \ok	& \ok	& \ok	& \err	& \err	& \ok	&	12 \\
\hline
\end{tabular}
\caption{\small Solver application to noisy data, $N=50$, $\sigma = 0.001$: Success (\ok\small) and failures (\err\small) for different applying each of 19 different solvers to each of 20 distinct data sets. The last column displays the number of successful computations per solver.}\label{fig:results.noise50}
\end{center}\end{minipage}
\end{tabular}
\end{table}

\subsection{Application to noisy data}\label{sec:experiments.noise}

Finally, we investigated the effect of noise in the data. We used the same data sets as above, but considered only those solvers and datasets for which the computation did not fail in the previous experiment. A random vector of independent zero-mean Gaussian components was added to each $\hat{X}_{i}$ and $\hat{Y}_{i}$, and the polynomial system was derived from this noisy data.

Our experiments revealed that even for a very small noise standard deviation, the computation completely failed in many cases. Here, failing means that though a reduced Gröbner basis was computed, no real valued solution could be obtained from that because the reduced basis contained equations like $0 = c_1h^2_k+c_2$ with positive coefficients $c_1$ and $c_2$. Selected results are illustrated in table~\ref{fig:results.noise}. For larger numbers of segments the failure percentage as well as the mean error decreases, such that on some data sets none of the generated solver fails for $N\geq 50$ as illustrated in table~\ref{fig:results.noise50}.

\section{Conclusion}\label{sec:conclusion}

In this paper, we proposed an algebraic way to compute a homography to upgrade a preliminary projective reconstruction to an Euclidean one by means of constraints derived from a minimal number of segments with known lengths. We believe, this could be useful in environments were prior camera calibration is impracticable but a small number of distances between points is known. 

We have shown that it is possible to solve this problem using only 9 segments and demonstrated how a corresponding solver can be constructed from simplified template data. However, our experiments revealed that the presented method is very likely to fail on noisy data. This has to be investigated more thoroughly in order to be able to apply our technique to real world data.

\section*{Acknowledgements}
This work has been supported by the project PRoViDE EU FP7-SPACE-312377 and FP7-SME-2011-285839 De-Montes.

\section*{Appendix: Notation}\label{sec:notation}
We use the notations \itshape{term}\upshape\ and \itshape{monomial}\upshape\ as they are explained in~\cite{Cox.etal_IdealsVarietiesand_1992}, i.e.\ given a polynomial ring $K[x_1,x_2,\ldots,x_n]$, a monomial is a product of the form 
\begin{equation*}
	x_1^{\alpha_1}\cdot x_2^{\alpha_2}\cdots x_n^{\alpha_n} = x^{\alpha}, 
\end{equation*}
with non-negative integer exponents $\alpha_1,\alpha_2,\ldots \alpha_n$. A term then denotes the product $a_{\alpha}x^{\alpha}$ of a monomial and a non-zero coefficient $a_{\alpha} \in K$.

An \itshape{$S$-polynomial}\upshape\ of a pair of polynomials $(f,g)$ is computed as 
\begin{equation*}
	S(f,g) = x^{\gamma}(\mbox{LT}(f))^{-1}f - x^{\gamma}(\mbox{LT}(g)^{-1}g 
\end{equation*} 
where $x^{\gamma} = \mbox{LCM}(\mbox{LM}(f),\mbox{LM}(g))$ is the least common multiple of $\mbox{LM}(f)$ and $\mbox{LM}(g)$. $\mbox{LM}(f)$ denotes the leading monomial and $\mbox{LT}(f)$ the leading term of $f$ w.r.t.\ a monomial ordering.

\bibliography{oagm2013}

\end{document}